# Forecasting Volatility in Indian Stock Market using Artificial Neural Network with Multiple Inputs and Outputs

Tamal Datta Chaudhuri
Principal, Calcutta Business School,
Diamond Harbour Road, Bishnupur – 743503, 24
Paraganas (South), West Bengal

Indranil Ghosh
Assistant Professor, Calcutta Business School,
Diamond Harbour Road, Bishnupur – 743503, 24
Paraganas (South), West Bengal

## ABSTRACT
Volatility in stock markets has been extensively studied in the applied finance literature. In this paper, Artificial Neural Network models based on various back propagation algorithms have been constructed to predict volatility in the Indian stock market through volatility of NIFTY returns and volatility of gold returns. This model considers India VIX, CBOE VIX, volatility of crude oil returns (CRUDESDR), volatility of DJIA returns (DJIASDR), volatility of DAX returns (DAXSDR), volatility of Hang Seng returns (HANGSDR) and volatility of Nikkei returns (NIKKEISDR) as predictor variables. Three sets of experiments have been performed over three time periods to judge the effectiveness of the approach.

## 1. INTRODUCTION
Volatility in stock markets evokes varying responses from market participants. While some perceive it as opportunity to make money, others perceive it as a threat and start unwinding their positions. While affecting portfolio choice, changes in stock market volatility also gives some idea about the current economic state. In today's globalized environment, increased volatility reflects global uncertainty. Volatility in the stock market as a whole can be due to macroeconomic factors, both internal and external. Examples could be oil price shocks, or increase in rates of interest in the US, or domestic elections. Volatility in individual stocks, on the other hand, can be due to perceived growth prospects of the company or the sector. It could also be triggered by company specific news or policy announcements.

The effects of stock market volatility can be sixfold. First, it enhances the profit making opportunities from intraday trading for spot market traders. Second, it leads to portfolio rebalancing by fund managers. Third, it increases volatility trading in the options market. Fourth, it increases hedging activity in financial markets. Fifth, it does influence policy makers in taking hard decisions as their actions can cause loss of wealth to retail holders. Sixth, it affects capital formation, as volatile markets are not conducive for fresh equity issues in the market.

While the effects of unanticipated announcements by companies, or external macroeconomic events like sovereign defaults, or economy wide policy changes on market volatility cannot be estimated, under normal market conditions, the Black and Scholes options pricing model provides a framework to estimate future volatility. This is denoted by "implied volatility", volatility that is expected to prevail in the near future as implied by the option price. In the spot market, if the expectation is that spot prices are going to fall, players rush to the options market to hedge their positions thus increasing implied volatility.

In the options pricing formula, the options price C

$C = f(S, K, t, \sigma, r)$

where S is the spot price of the underlying, K is the strike price, t is the time to expiry, $\sigma$ is volatility and r is the rate of interest. In the options market, the players cannot influence S, t or r. K they have to choose themselves. The only two variables that remain are C and $\sigma$. If we substitute the value of historic volatility in place of $\sigma$, then we will solve for the theoretical options price. If we plug in the value of the actual traded price of the options contract, then we will solve for implied volatility. The latter is an estimate of the actual volatility that is expected to prevail in the next three to four weeks. Thus, actual volatility and implied volatility should move together. The various possible movements between historic volatility and implied volatility has been described in detail in Passarelli (2008).

In todays globalized environment, with increased financial integration and also enhanced trade in goods and services, volatility in one country spreads to other countries almost immediately. In India, where foreign institutional investors (FIIs) are large players in the stock market, their fund allocation is shaped by macroeconomic conditions in other economies. Thus any macroeconomic event in any part of the world causes reallocation of FII funds, leading to volatility in Indian stock markets.

Generally, when stock market becomes volatile, there is a tendency for gold prices to rise. It is considered to be a safe asset and hence there is a tendency to substitute stocks with gold. Thus volatility in gold prices is also a reflection of volatility in stock markets.

The purpose of this paper is to develop a framework for forecasting volatility in the Indian stock market.

## 2. OBJECTIVE OF THE STUDY

The paper proposes an Artificial Neural Network (ANN) framework for forecasting volatility in the Indian stock market. The model has volatility of NIFTY returns and volatility of gold returns as the two outputs. It has India VIX, CBOE VIX, volatility of crude oil returns, volatility of DJIA returns, volatility of DAX returns, volatility of Hang Seng returns and volatility of Nikkei returns as the seven inputs. The objective is to capture the effects of both external and internal shocks on spot market volatility. The advantage of using the ANN framework is that it does not presuppose any linearity in the relationship between the inputs and outputs. Further, it allows for interaction and feedback between the inputs. We do not use lagged values of the outputs as inputs to avoid time dependency and we model external shocks through crude oil market volatility as well as volatility in other financial markets. Internal shocks are assumed to be represented through movements in India VIX.

Accordingly, the plan of the paper is as follows. The ANN framework for our study is described in Section 3. A literature survey is presented in Section 4. The choices of variables are discussed in Section 5. The data and the results of the study are discussed in Section 6 and Section 7 concludes the paper.

## 3. METHODLOGY

Artificial Neural Networks (ANN) are effective machine learning tools, that mimic the working nature of the human brain, in order to identify the associative pattern between a set of inputs and outputs. Human brain is a massively interconnected structure of around $10^{10}$ number of basic processing units known as neurons. Similar to this architecture, in ANN, neurons are structured and connected in a hierarchical manner. A distinct input layer and output layer are interlinked (artificial synapses) through a single or multiple hidden layer(s). Strength of each connection between any two neurons is represented by numeric weight value. These weight values actually correspond to the decision boundary obtained by the ANN classifier. When a given set of input and output values of variables under study are presented to a Neural Network as training dataset, weight values are estimated via different learning algorithms. Once the estimated values are stabilized after validation, trained ANN is tested against a test data set to evaluate its predictive power.

Figure 1 depicts a typical ANN architecture with five inputs, one hidden layer and one output.

**Figure 1: A simple ANN model**

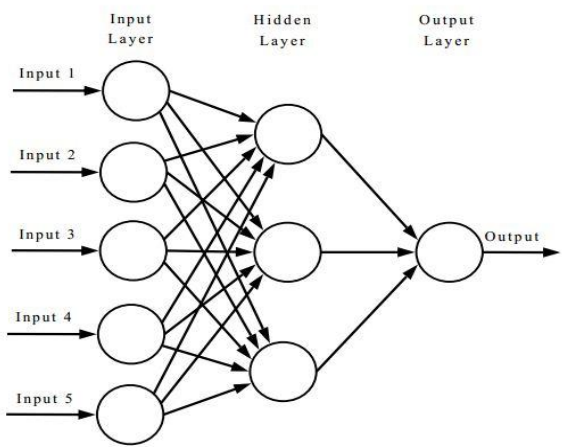

*Source: Authors' own construction*

Strength of each connection between any two neurons is represented by numeric weight value. These weight values actually correspond to the decision boundary obtained by ANN classifier. When a given set of input and output values of variables under study are presented to a Neural Network as training dataset, weight values are estimated via different learning algorithms. Once the estimated values are stabilized after validation, trained ANN is tested against a test data set to evaluate its predictive power. Each input signal ($x_i$) is associated with a weight ($w_i$). The overall input I to the processing unit is a function of all weighted inputs given by.

$$I = f(\sum x_i \times w_i) \quad (1)$$

The activation state of the processing unit (A) at any time is a function (usually nonlinear) of I

$$A = g(I) \quad (2)$$

The output Y from the processing unit is determined by the transfer function h

$$Y = h(A) = h(g(I)) = h(g(f(\sum x_i \times w_i)$$
$$= \Theta(\sum x_i \times w_i) \quad (3)$$

An ANN is said to "learn" mapping for a function or a process. Since the topology, the activation function A, and the transfer function h are normally fixed at the time the network is constructed, the only adjustable parameters are the weights $w_i$. Learning means changing the weights adaptively to meet some criterion based on the signals from the output units (nodes). A common training algorithm for ANN is back propagation (a steepest gradient descent method). It minimizes the sum of the squares of the differences between vectors Y and $Y_d$ .i.e.,

$$E = \frac{1}{2}(Y - Y_d)^T (Y - Y_d) \quad (4)$$

Where Y represents a vector of outputs of all the output nodes, $Y_d$ is a vector of desired outputs, and superscript T stands for standard transpose operation. Many types of ANN models have been proposed during the last two decades to map inputs to outputs. Among them, layered ANN's trained by a back-propagation learning algorithm forms the basis of most common practical applications. Weight and bias matrix associated with the inputs are adjusted/updated by using some learning rule or training algorithm which is non-linear. Based on the general relations in (1) through (3)**,** the outputs from the input layer to the hidden layer and the outputs from the hidden layer to the output layer of the network are, respectively,

$$Z = \Theta_z(Z\,\Omega_z) \quad (5)$$

and

$$Y = \Theta_y(Z\,\Omega_y) \quad (6)$$

Where $\Theta_z$, and $\Theta_y$, are usually sigmoid functions which can be described by the following expression

$$O_j = 1/(1 + \exp(i_j)) \quad (7)$$

Where $O_j$ is the output of node j and $i_j$ is the net-input of node j.

Due to its efficacy in parallel processing to mine complex nonlinear patterns, ANN has garnered a lot of attention in pattern recognition literature. Ability to operate in nonparametric environment has given it competitive edge over traditional statistical tools such as regression analysis. It has been highly successful both in predicting the state of categorical variable(s) and forecasting the outcome of continuous variable(s) as cited in literature. Complex real world problems such as prediction of financial health, bankruptcy prediction, stock index return analysis, credit default analysis, manufacturing assembly line balancing, PID controller monitoring, Enterprise Resource Planning performance analysis, etc. have been analyzed using the ANN framework.

## 4. LITERATURE REVIEW

In this section, we present the two strands of the literature on which this paper is based. The first is application of ANN in various areas of research which reflects the wide range applicability of this tool of analysis. The second is research papers on forecasting volatility in stock markets including those which have applied ANN as a tool of analysis.

Walczak and Sincich (1999) made a comparative analysis of neural network and logistic regression in student profile selection for university enrolments and the results showed that ANN outperformed logistic regression. Ling and Liu (2004) investigated the critical success factors of design-build projects in Singapore through ANN based modelling where eleven success measures and sixty five factors were analyzed. Karnik et al. (2008) utilized multilayer feed forward ANN trained by backpropagation algorithm to model and critically examine the impact of drilling process parameters on the delamination factor. Pal et al. (2008) designed a multilayer ANN model to estimate the tensile stress of welded plates and compared the results with multiple regression analysis. Rouhani and Ravasan (2012) investigated the relationship between organizational factors and post Enterprise Resource Planning System implementation success using a novel Neural Network framework. Ndaliman et al. (2012) proposed an ANN model with multi-layer perception neural architecture for the prediction of SR on first commenced Ti-15-3 alloy in electrical discharge machining (EDM) process. Zhao et al. (2015) utilized wavelet neural network and proposed a variable step size updating learning algorithm for parameter tuning operation of PID controller. Ramasamy et al. (2015) attempted to predict wind speeds of different locations (Bilaspur, Chamba, Kangra, Kinnaur, Kullu, Keylong, Mandi, Shimla, Sirmaur, Solan and Una location) in the Western Himalayan Indian state of Himachal Pradesh adopting ANN based framework. Ghiassi et al. (2015) applied dynamic artificial neural network to forecast movie revenues during the pre-production period in USA using MPAA rating, sequel, number of screens, production budgets, pre-release advertising expenditures, runtime & seasonality as predictor variables. Oko et al. (2015) presented a dynamic model of the drum-boiler to predict drum pressure and level in coal-fired subcritical power plant using NARX neural networks. Aish et al. (2015) incorporated Multilayer perceptron (MLP) and radial basis function (RBF) neural networks as prediction tool to forecast reverse osmosis desalination plant's performance in the Gaza Strip.

In the second strand of the literature, Rather et al. (2015) employed to two linear models namely auto regressive moving average and exponential smoothing and recurrent neural network as a nonlinear model to predict returns of six stocks (TCS, BHEL, Wipro, Axis Bank, Maruti & Tata Steel) using training dataset from National Stock Exchange of India (NSE). Results showed the supremacy of neural model over the linear models. Further, authors proposed a hybrid prediction model that use the results of individual prediction models and tested the effectiveness of it in estimating returns from twenty five stocks belonging to different industrial sectors.

Adhikary (2015) presented an ANN based ensemble prediction framework for time series forecasting problems. Malliaris and Salchenberger (1996) employed Elman's recurrent neural network and ARIMA model in forecasting copper spot prices using New York Commodity Exchange (COMEX) data. The study reports that the neural model outperforms ARIMA model in terms of forecasting accuracy.

Malhotra (2012) attempted to examine the impact of stock market futures on spot market volatility for selected stocks from key industry sectors. The GARCH technique was used to capture the time varying nature of volatility of the Indian stock market. Tripathy and Rahman (2013) also use the GARCH model for forecasting daily stock volatility. Vegendla and Enke (2013) investigate the forecasting ability of Feedback Forward Neural Network using back propagation learning Recurrent Neural Networks and also GARCH models of historic volatility, implied volatility and model based volatility. The exercise is done for NASDAQ, DJIA, NYSE and S & P 500.

Panda and Deo (2014), Srinivasan and Prakasham (2014) and Srinivasan (2015), using different sets of variables, apply the GARCH model or the Autoregressive Distributed Lag model, to understand the volatility spillover between various financial assets.

In a recent contribution, Dixit, Roy and Uppal (2013) have provided a framework for predicting India VIX using Artificial Neural Network. In their model, the first seven indicators are current day's open (CO), high (CH), low (CL) and close (CC) index values followed by previous day's high (PH), low (PL) and close (PC) index values. Next four input parameters were calculated using the simple moving average of the last (including the current day) 3 days (SMA3), 5 days (SMA5), 10 days (SMA10) and 15 days (SMA15) closing India VIX values.

McMillan (2004) presents a non-parametric framework for predicting implied volatility where Implied Volatility (IV) and Historic Volatility (HV) are grouped in deciles. This is discussed in detail in Datta Chaudhuri and Sheth (2014) where such deciles are constructed for India VIX (IV) and standard deviation of NIFTY returns (HV). The methodology involves taking a 20 Day Moving Average (MA) of IV and a 10 Day, 20 Day and 50 Day Moving Averages of HV up to a date and constructing deciles. Then the actual values of the variables are computed on a subsequent date, outside the cut-off date, and the decile position of the values is marked off. This information is then used to execute options trading strategies.

## 5. THE VARIABLES

Together with our methodology, our paper differs from the existing literature in the choice of inputs and outputs. We do not take lagged values of volatility as the inputs. Further, we allow for two outputs namely volatility of NIFTY returns (NIFTYSDR) and volatility of gold returns (GOLDSDR). To calculate NIFTYSDR, we take 20 day rolling standard deviation, annualized, of NIFTY returns. This is historic volatility and this is one of the variables that we want to predict. The other output is GOLDSDR which is also calculated as 20 day rolling standard deviation of gold returns, annualized.

As inputs we consider India VIX, CBOE VIX, volatility of crude oil returns (CRUDESDR), volatility of DJIA returns

(DJIASDR), volatility of DAX returns (DAXSDR), volatility of Hang Seng returns (HANGSDR) and volatility of Nikkei returns (NIKKEISDR). As discussed earlier, INDIA VIX, as derived from the options market, is a forward looking indicator for actual volatility. So it finds place in our analysis as a predictor or input. We do not explicitly consider lagged values of NIFTYSDR as inputs, as the ANN framework would consider feedback from past values. Further, it would also allow for learning from future values. To allow for external shocks, as India is a large importer of crude oil, we consider CRUDESDR an input. In the recent past, political instability in the Middle East and related regions have impacted the expected availability of oil and has resulted in stock market instability in India. Global macroeconomic impacts have been incorporated through DJIASDR, DAXSDR, HANGSDR and NIKKEISDR. We have considered the impact of instability in both the western world and the eastern world. The inclusion of CBOE VIX is to factor in the impact of expected future volatility in the US market on the Indian market. That is, if CBOE VIX rises, some future instability in the US markets is foreseen. This in turn affects FII fund flows and hence NIFTYSDR.

**Figure 2: INDIAVIX and NIFTYSDR for the period 3.3.2008 to 10.4.2015**

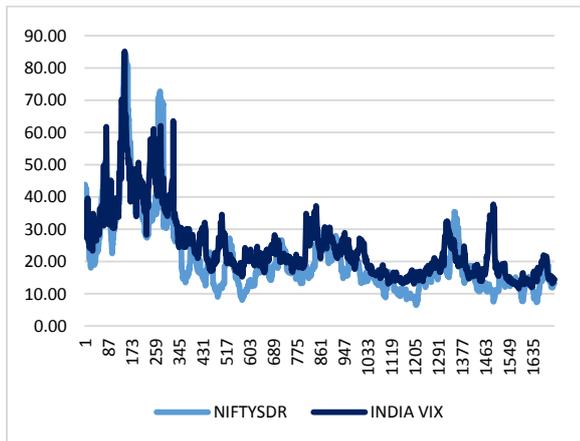

*Source: Authors' own construction*

Figure 2 clearly suggests that, overall, over a fairly long period, historic volatility and implied volatility do move together. So considering INDIA VIX as a predictor of NIFTYSDR is alright.

The following Figures 3 and 4 show the movement in the two variables in different sub periods.

**Figure 3: INDIAVIX and NIFTYSDR for 2013**

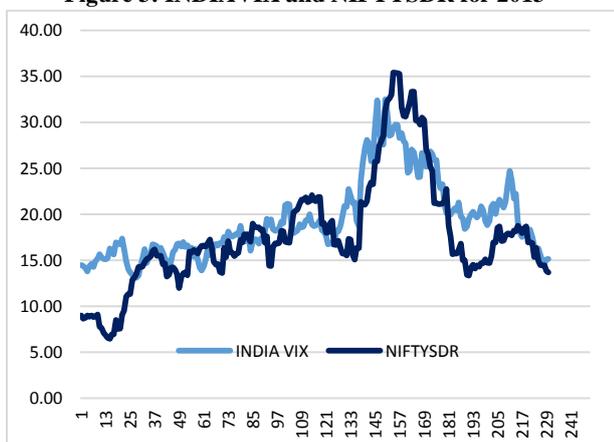

*Source: Authors' own construction*

Figures 3 and 4 reveal that for shorter time periods, the movements in the two variables are not always in tandem and hence the rationale for inclusion of other inputs in the analysis.

**Figure 4: INDIAVIX and NIFTYSDR for 2014**

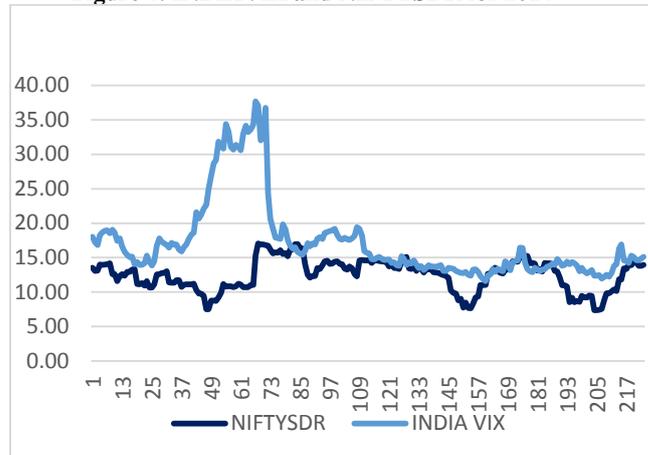

*Source: Authors' own construction*

**Figure 5: India VIX for the period 5.3.2008 – 21.4.2015**

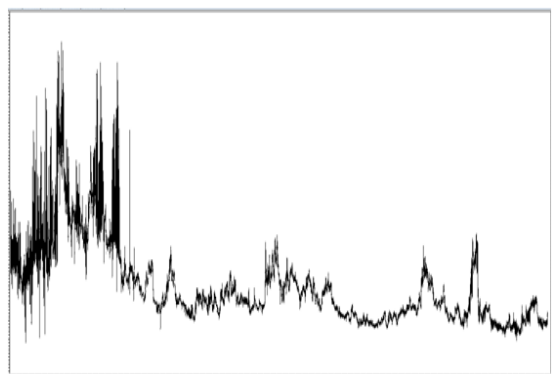

*Source: Metastock*

Figure 5 depicts the impact of global financial crisis of 2008 on INIDA VIX and clearly there are global factors that enter domestic expectations formation. This becomes even clear from Figure 6 where expected volatility in the US seems to go hand in hand with expected volatility in India. That is, global uncertainties affect US implied volatility, which in turn affects implied volatility index in India. There are, however, discrepancies, and hence both enter as inputs in our study.

**Figure 6: INDIA VIX and CBOE VIX 2008 onwards**

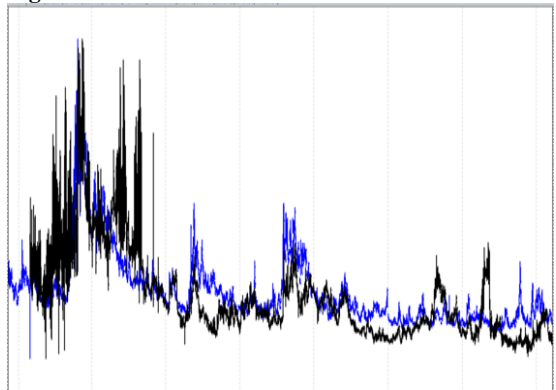

*Source: Metastock*

## 6. RESULTS AND ANALYSIS

In this paper, three experiments have been conducted. In the first experiment, attempt has been made to forecast NIFTY Returns and Gold Returns for first four months of 2015 utilizing the entire daily data on the variables for the years 2013 and 2014 as training data. In experiment two, data for the entire year 2013 and a major part of 2014 has been used as training data to predict NIFTY Returns and Gold Returns for a part of 2014. In the third experiment, the training data for the first experiment has been used to estimate NIFTY Returns and Gold Returns for a past period, year 2008. The latter has been done to examine whether the adopted framework can estimate market volatility of some past period based on the present scenario. This is our way of understanding the nature of the data and also the analytical framework used. It is also a means of validating our approach.

In this paper, two different neural architecture and nine learning algorithms have been adopted. One hidden layer is used while number of hidden neurons has been varied at three levels (20, 30 & 40 respectively). Hence total number of trials is fifty four (2*9*3). Descriptive statistics of different performance indicators are presented to judge the results critically. Other important specifications of parameters which have been used throughout the ANN modeling process are shown in Table 1.

**Table 1: Important specification of parameters**

| Sl. No. | Parameter | Data/Technique Used |
|---|---|---|
| 1. | Number of input neuron(s) | Seven |
| 2. | Number of output neuron(s) | Two |
| 3. | Transfer function(s) | Tan-sigmoid transfer function (tansig) in hidden layer &purelin in output layer. |
| 4. | Error function(s) | Mean squared error(MSE) function |
| 5. | Type of Learning rule | Supervised learning rule |

*Source: Authors' own construction*

To evaluate the performance of the framework, three metrics namely mean squared error (MSE), correlation between predicted and actual values (R) and mean absolute percentage error (MAPE) have been used. MSE is expressed as

$$\text{MSE} = \frac{1}{N}\sum_{i=1}^{N}\{Y_{act}(i) - Y_{pred}(i)\}^2$$

R is the correlation measure between the actual and predicted outcomes which is computed as

$$R = \frac{N(\sum_{i=1}^{N} Y_{act}(i) * Y_{pred}(i)) - (\sum_{i=1}^{N} Y_{act}(i)) * (\sum_{i=1}^{N} Y_{pred}(i))}{\sqrt{\left[N\sum_{i=1}^{N} Y_{act}(i)^2 - \left(\sum_{i=1}^{N} Y_{act}(i)\right)^2\right]\left[N\sum_{i=1}^{N} Y_{pred}(i)^2 - \left(\sum_{i=1}^{N} Y_{pred}(i)\right)^2\right]}}$$

MAPE is the average sum of absolute percentage error(s) over the entire dataset. Mathematically it is calculated as:

$$\text{MAPE} = \frac{1}{N}\sum_{i=1}^{N}\left|\frac{Y_{act}(i) - Y_{pred}(i)}{Y_{act}(i)}\right| \times 100\%$$

where N denotes the total number of observations.

Descriptive statistics of all three performance indicators for experiment 1 are shown in Tables 2 to 7.

**Table 2: MSE of total 54 trials for the training dataset**

| MSE | Multi-Layer Feed Forward Network | Cascade Feed Forward Network |
|---|---|---|
| Min | 1.0528 | 1.8691 |
| Max | 3.9826 | 4.0125 |
| Average | 2.4627 | 2.6853 |
| Standard Deviation | 1.0424 | 1.2507 |

*Source: Authors' own construction*

**Table 3: R of total 54 trials for the training datasets**

| R | Multi-Layer Feed Forward Network | Cascade Feed Forward Network |
|---|---|---|
| Min | 0.9427 | 0.9358 |
| Max | 0.9842 | 0.9742 |
| Average | 0.9643 | 0.9543 |
| Standard Deviation | 0.0142 | 0.0205 |

*Source: Authors' own construction*

**Table 4: MAPE of total 54 trials for the training datasets**

| MAPE | Multi-Layer Feed Forward Network | Cascade Feed Forward Network |
|---|---|---|
| Min | 0.7653 | 0.8003 |
| Max | 1.0592 | 1.1203 |
| Average | 0.9021 | 0.9207 |
| Standard Deviation | 0.1071 | 0.1103 |

*Source: Authors' own construction*

**Table 5: MSE of total 54 trials for the testing datasets**

| MSE | Multi-Layer Feed Forward Network | Cascade Feed Forward Network |
|---|---|---|
| Min | 3.0244 | 3.5648 |
| Max | 4.1282 | 4.3206 |
| Average | 3.7251 | 3.9473 |
| Standard Deviation | 0.3581 | 0.3948 |

*Source: Authors' own construction*

**Table 6: R of total 54 trials for the testing datasets**

| R | Multi-Layer Feed Forward Network | Cascade Feed Forward Network |
|---|---|---|
| Min | 0.9534 | 0.9431 |
| Max | 0.9748 | 0.9763 |
| Average | 0.9654 | 0.9647 |
| Standard Deviation | 0.0068 | 0.0093 |

*Source: Authors' own construction*

**Table 7: MAPE of total 54 trials for the testing datasets**

| MAPE | Multi-Layer Feed Forward Network | Cascade Feed Forward Network |
|---|---|---|
| Min | 0.8328 | 0.8773 |
| Max | 1.1726 | 1.2016 |
| Average | 0.9592 | 1.018 |
| Standard Deviation | 0.1227 | 0.1904 |

*Source: Authors' own construction*

MSE and MAPE values must be as low as possible to indicate efficient prediction; ideally a value of zero signifies no error. On the other hand, a value of R close to 1 is a must for strong prediction. For both training and test dataset, the values of the performance indicators shown above justify the effectiveness of MLFF and CFFN tool as a forecasting tool for the problem at hand. It can thus be concluded that volatility of NIFTY Returns and Gold Returns can be predicted using India VIX, CBOE VIX, CRUDESDR, DJIASDR, DAXSDR, HANGSDR and NIKKEISDR.

Figure 7 depicts the regression plot of forecast values as generated using the test data as against the actual data of 2015. The results indicate that the methodology used and the inputs chosen forecast the volatility of the outputs well.

**Figure 7: Regression plot of Experiment 1.**

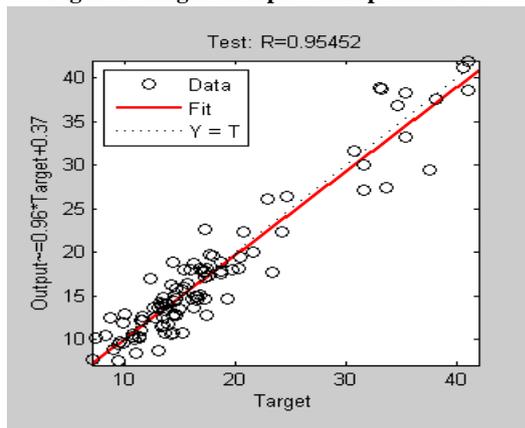

*Source: Matlab*

For Experiment 2, we have kept the same experimental settings and the results are displayed in the following tables.

**Table 8: MSE of total 54 trials for the training datasets**

| MSE | Multi-Layer Feed Forward Network | Cascade Feed Forward Network |
|---|---|---|
| Min | 2.8214 | 2.9403 |
| Max | 4.3251 | 3.8923 |
| Average | 3.6284 | 3.5981 |
| Standard Deviation | 0.5065 | 0.4818 |

*Source: Authors' own construction*

**Table 9: R of total 54 trials for the training datasets**

| R | Multi-Layer Feed Forward Network | Cascade Feed Forward Network |
|---|---|---|
| Min | 0.9432 | 0.9581 |
| Max | 0.9872 | 0.9868 |
| Average | 0.9604 | 0.9677 |
| Standard Deviation | 0.0158 | 0.0139 |

*Source: Authors' own construction*

**Table 10: Statistics of MAPE of total 54 trials for the training datasets**

| MAPE | Multi-Layer Feed Forward Network | Cascade Feed Forward Network |
|---|---|---|
| Min | 0.8603 | 0.8018 |
| Max | 1.1209 | 1.0962 |
| Average | 1.0062 | 0.9623 |
| Standard Deviation | 0.0935 | 0.0968 |

*Source: Authors' own construction*

**Table 11: MSE of total 54 trials for the testing datasets**

| MSE | Multi-Layer Feed Forward Network | Cascade Feed Forward Network |
|---|---|---|
| Min | 3.1267 | 3.2207 |
| Max | 4.2582 | 4.1263 |
| Average | 3.7508 | 3.5262 |
| Standard Deviation | 0.4236 | 0.3708 |

*Source: Authors' own construction*

**Table 12: R of total 54 trials for the testing datasets**

| R | Multi-Layer Feed Forward Network | Cascade Feed Forward Network |
|---|---|---|
| Min | 0.9268 | 0.9325 |
| Max | 0.9634 | 0.9662 |
| Average | 0.9483 | 0.9518 |
| Standard Deviation | 0.0143 | 0.0127 |

*Source: Authors' own construction*

**Table 13: MAPE of total 54 trials for the testing datasets**

| MAPE | Multi-Layer Feed Forward Network | Cascade Feed Forward Network |
|---|---|---|
| Min | 0.8457 | 0.8223 |
| Max | 1.1063 | 1.0218 |
| Average | 0.9641 | 0.9414 |
| Standard Deviation | 0.0906 | 0.0892 |

*Source: Authors' own construction*

Figure 8 again indicates that for the second experiment also the methodology used and the inputs chosen forecast the volatility of the outputs well.

**Figure 8: Regression plot of Experiment 2.**

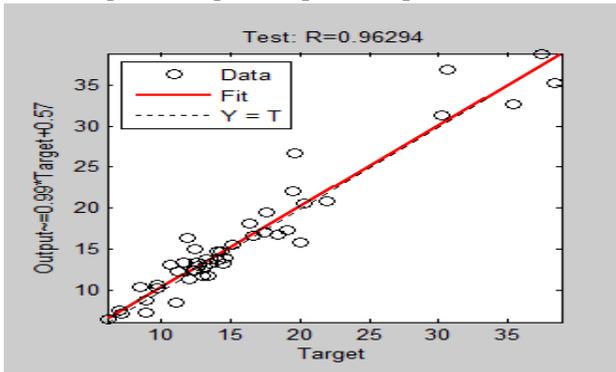

*Source: Matlab*

The third experiment that we perform is interesting as we have been employed the data for 2013 and 2014 together as training data to estimate the volatility back in 2008. Tables 14-19 portray the results and the findings are discussed later.

**Table 14: MSE of total 54 trials for the training datasets**

| MSE | Multi-Layer Feed Forward Network | Cascade Feed Forward Network |
|---|---|---|
| Min | 14.2362 | 15.6243 |
| Max | 22.3898 | 23.1684 |
| Average | 19.7424 | 20.1871 |
| Standard Deviation | 2.3516 | 2.2783 |

**Source: Authors' own construction**

**Table 15: R of total 54 trials for the training datasets**

| R | Multi-Layer Feed Forward Network | Cascade Feed Forward Network |
|---|---|---|
| Min | 0.8891 | 0.8934 |
| Max | 0.9251 | 0.9362 |
| Average | 0.9054 | 0.9126 |
| Standard Deviation | 0.0107 | 0.0125 |

*Source: Authors' own construction*

**Table 16: MAPE of total 54 trials for the training datasets**

| MAPE | Multi-Layer Feed Forward Network | Cascade Feed Forward Network |
|---|---|---|
| Min | 5.7682 | 5.5803 |
| Max | 8.0974 | 7.9561 |
| Average | 6.6785 | 6.4327 |
| Standard Deviation | 0.5983 | 0.5204 |

*Source: Authors' own construction*

**Table 17: MSE of total 54 trials for the testing datasets**

| MSE | Multi-Layer Feed Forward Network | Cascade Feed Forward Network |
|---|---|---|
| Min | 16.2218 | 15.9583 |
| Max | 22.3735 | 22.5612 |
| Average | 18.8187 | 19.0462 |
| Standard Deviation | 2.0846 | 2.1283 |

*Source: Authors' own construction*

**Table 18: R of total 54 trials for the testing datasets**

| R | Multi-Layer Feed Forward Network | Cascade Feed Forward Network |
|---|---|---|
| Min | 0.8772 | 0.8806 |
| Max | 0.9184 | 0.9189 |
| Average | 0.8923 | 0.8918 |
| Standard Deviation | 0.0138 | 0.0126 |

*Source: Authors' own construction*

**Table 19: MAPE of total 54 trials for the testing datasets**

| MAPE | Multi-Layer Feed Forward Network | Cascade Feed Forward Network |
|---|---|---|
| Min | 6.0182 | 6.0143 |
| Max | 8.3264 | 8.7065 |
| Average | 7.2165 | 7.5781 |
| Standard Deviation | 0.77 | 0.7981 |

*Source: Authors' own construction*

Interestingly in this experiment, it can be seen that the average MSE and MAPE values for both training and testing set are considerably larger in compared to the earlier experiments. Similarly average R values are also lower in both training and testing set. So it may be inferred that prediction accuracy of the model trained in present time has goes down when asked to forecast market volatility back in 2008. Given the extent of enormously increased volatility in 2008 post the crisis, as shown in Figure 5, the results are not very surprising. The regression plot in Figure 9 also captures this.

**Figure 9 Regression plot of Experiment 3.**

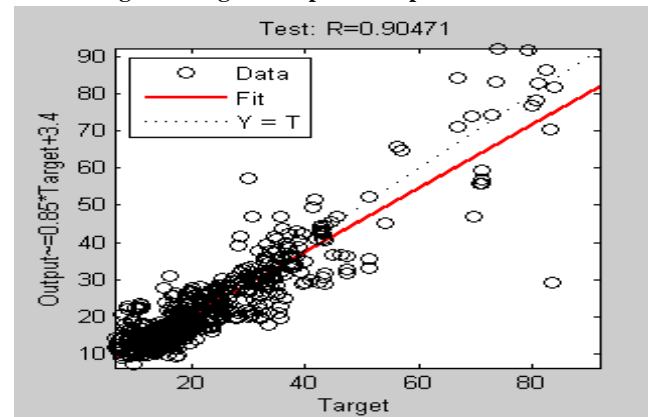

*Source: MATLAB*

In Figures 10, 11 and 12, we portray the performance of the two different neural architectures on test data set.

A comparative analysis of MLFF and CFFN has also been carried out to statistically analyze their performance. Statistical t-test has been conducted on MSE to judge whether their performances are significantly different or not. Table 20 depicts the outcomes.

**Table 20: Significance values (on test cases)**

| Experiment 1 | Experiment 2 | Experiment 3 |
|---|---|---|
| 0.221 (two tailed) | 0.305 (two tailed) | 0.184 (two tailed) |

*Source: Authors' own construction*

As none of the values of the test statistic are significant, it can be concluded that there is no significant difference in performance among two models.

**Figure 10: EXPERIMENT 1**

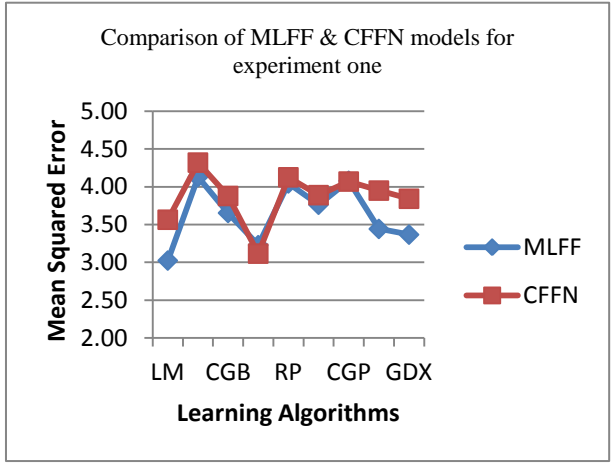

*Source: Authors' own construction*

**Figure 11: EXPERIMENT 2**

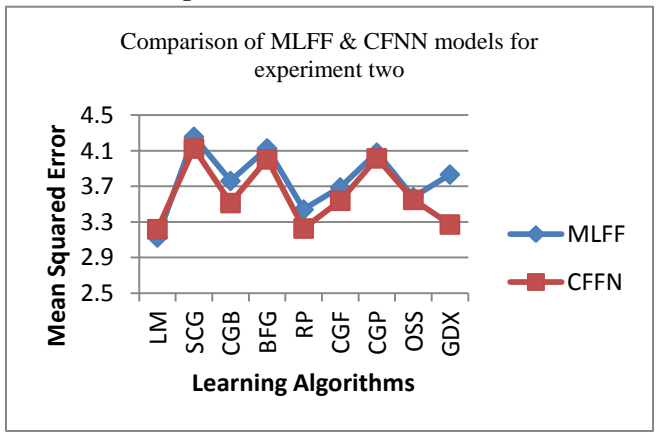

**Source: Authors' own construction**

**Figure 11: EXPERIMENT 3**

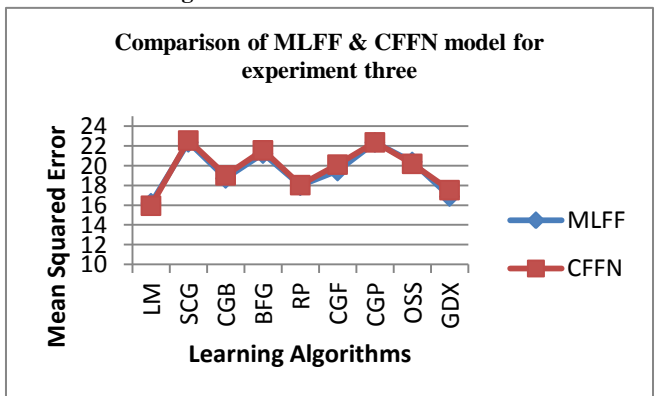

*Source: Authors' own construction*

## 7. CONCLUSION

The purpose of this paper was to examine the efficacy of the ANN framework in predicting volatility in the Indian stock market. We used a multiple input multiple output structure using two different neural architecture and nine learning algorithms. For our experiments, only one hidden layer was used while number of hidden neurons has been varied at three levels (20, 30 & 40 respectively). Hence total number of trials was fifty four (2*9*3). We conducted our exercise for three different time periods. Our framework could satisfactorily forecast volatility for 2015 using training data for 2013-14. However, the prediction accuracy of the model, trained in present time, has gone down when asked to forecast market volatility back in 2008.

## 8. REFERENCES

[1] Adhikari, R., (2015), A neural network based linear ensemble framework for time series forecasting, Neurocomputing, 157, 231-242

[2] Aish, A. M., Zaqoot H. A. & Abdeljawad, S. M., (2015), Artificial neural network approach for predicting reverse osmosis desalination plants performance in the Gaza Strip, Desalination, 367, 240-247.

**[3]** Datta Chaudhuri, T and Kinjal, S., (2014), Forecasting Volatility, Volatility Trading and Decomposition by Greeks, CBS Journal of Management Practices, 1, 59-70.

[4] Dixit, G., Roy, D. & Uppal, N., (2013), Predicting India Volatility Index: An Application of Artificial Neural Network, 70, 22-30.

[5] Ghiassi, M., Lio, D. & Moon, B., (2015), Pre-production forecasting of movie revenues with a dynamic artificial neural network, Expert Systems with Applications, 42, 3176-3193.

[6] Karnik, S. R., Gaitonde, V. N., Campos Rubio, J., EstevesCorreia, A., Abrão, A. M., & Paulo Davim, J. (2008). Delamination analysis in high speed drilling of carbon fiber reinforced plastics (CFRP) using artificial neural network model, Materials and Design, 29, 1768–1776.

[7] Lasheras, F. S., Juez, de cos Juez, F. J., Sanchez, A. S., Krzemie, A. & Fernandez, P. R., (2015), Forecasting the COMEX copper spot price by means of neural networks and ARIMA models, Research Policy, 45, 37-43.

[8] Malhotra, G., (2012), Impact of Futures Trading on Volatility of CNX Nifty, IIMS Journal of Management Science, 3, 166-178.

[9] Malliaris, M. & Salchenberger, L., (1996), Using neural networks to forecast the S &P 100 implied volatility, Neurocomputing, 10, 183-195.

[10] McMillan, Lawrence G (2004), McMillan on Options, John Wiley & Sons, Inc., Hoboken, New Jersey.

[11] Ndaliman, M. B., Hazza, M., Khan, A. A. & Ali. M. Y., (2012) Development of a new model for predicting EDM properties of Cu-TaC compact electrodes based on artificial neural network method, Australian Journal of Basic and Applied Sciences, 6, 192-199.

[12] Oko, E., Meihong, W. & Zhang, J., (2015), Neural network approach for predicting drum pressure and level in coal-fired subcritical power plant, Fuel, 151, 139-145.

[13] Pal, M., Pal, S. K. & Samantaray, A. K., (2008), Artificial neural network modeling of weld joint strength prediction of a pulsed metal inert gas welding process using arc signals, Journal of Materials Processing Technology, 202, 464-474.

[14] Panda, P. & Deo, M., (2014) Asymmetric and Volatility Spillover Between Stock Market and Foreign Exchange Market: Indian Experience, IUP Journal of Applied Finance, 20, 69-82.

[15] Passarelli, D., (2008), Trading Options Greeks, Bloomberg Press, New York.

[16] Ramasamy, P., Chandel, S. S. & Yadav, A. K., (2015), Wind speed prediction in the mountainous region of India using an artificial neural network model, Renewable Energy, 80, 338-347.

[17] Rather, A., M., Agarwal, A. & Sastry, V. N., (2015), Recurrent neural network and a hybrid model for prediction of stock returns, Expert Systems with Applications, 42, 3234-3241.

[18] Srinivasan, P.& Karthigen, P., (2014), Gold Price, Stock Price and Exchange Rate Nexus: The Case of India, IUP Journal of Financial risk management, 11, 52-62.

[19] Srinivasan, P. (2015), Modelling and Forecasting Time-Varying Conditional Volatility of Indian Stock Market, IUP Journal of Financial Risk management, 12, 49-64.

[20] Tripathy., S.& Rahman, A.(2013), Forecasting Daily Stock Volatility Using Garch Model: A Comparison Between BSE and NSE, IUP Journal of Applied Finance, 15, 71-83.

[21] Vejendla, A. &Enke, D. (2013), Evaluation of GARCH, RNN & FNN Models for Forecasting Volatility in the Financial Markets, IUP Journal of Financial Risk management, 10, 41-49.

[22] Walczak, S. & Sincich, T. (1999), A Comparative Analysis of Regression and Neural Networks for University Admissions. Information Sciences, 119, 1-20.

[23] Zhao, Y., Du, X., Xia, G. & Wu., L., (2015), A novel algorithm for wavelet neural network s with application to enhanced PID controller designs, Neurocomputing, 158, 257-267.